\documentclass[conference]{IEEEtran}
\IEEEoverridecommandlockouts
\usepackage[noadjust]{cite}
\usepackage{caption}
\usepackage{amsfonts}
\usepackage{amsmath}
\usepackage{algorithm}
\usepackage[noend]{algpseudocode}
\usepackage{hyperref}
\usepackage{pgfplots}
\pgfplotsset{compat=1.17}
\usepackage{multirow}
\usepackage{subcaption}
\usepackage{mathtools}
\usepackage{tikz}
\usepackage{xcolor}
\usepackage[most]{tcolorbox}
\usepackage{makecell}
\usepackage{trimclip}

\usetikzlibrary{patterns}

\newcommand\compactdots{\hbox to 0.7em{.\hss.\hss.}}

\usepackage{newfloat}
\usepackage{listings}
\DeclareCaptionStyle{ruled}{labelfont=normalfont,labelsep=colon,strut=off} 
\floatstyle{ruled}
\newfloat{listing}{tb}{lst}{}
\floatname{listing}{Listing}
\def\BibTeX{{\rm B\kern-.05em{\sc i\kern-.025em b}\kern-.08em
    T\kern-.1667em\lower.7ex\hbox{E}\kern-.125emX}}
\begin{document}

\title{Concept Visualization: Explaining the CLIP Multi-modal Embedding Using WordNet

}

\author{\IEEEauthorblockN{Loris Giulivi, Giacomo Boracchi}
\IEEEauthorblockA{\textit{DEIB}, Politecnico di Milano \\
\{name.surname\}@polimi.it}
}
\maketitle

\begin{abstract}
Advances in multi-modal embeddings, and in particular CLIP, have recently driven several breakthroughs in Computer Vision (CV). CLIP has shown impressive performance on a variety of tasks, yet, its inherently opaque architecture may hinder the application of models employing CLIP as backbone, especially in fields where trust and model explainability are imperative, such as in the medical domain.
Current explanation methodologies for CV models rely on Saliency Maps computed through gradient analysis or input perturbation. However, these Saliency Maps can only be computed to explain classes relevant to the end task, often smaller in scope than the backbone training classes.
In the context of models implementing CLIP as their vision backbone, a substantial portion of the information embedded within the learned representations is thus left unexplained.

In this work, we propose Concept Visualization (ConVis), a novel saliency methodology that explains the CLIP embedding of an image by exploiting the multi-modal nature of the embeddings. ConVis makes use of lexical information from WordNet to compute task-agnostic Saliency Maps for any concept, not limited to concepts the end model was trained on.
We validate our use of WordNet via an out of distribution detection experiment, and test ConVis on an object localization benchmark, showing that Concept Visualizations correctly identify and localize the image's semantic content. Additionally, we perform a user study demonstrating that our methodology can give users insight on the model's functioning.


\end{abstract}

\begin{figure*}
    \centering
    \resizebox{1.0\textwidth}{!}{
        \includegraphics[width=\linewidth]{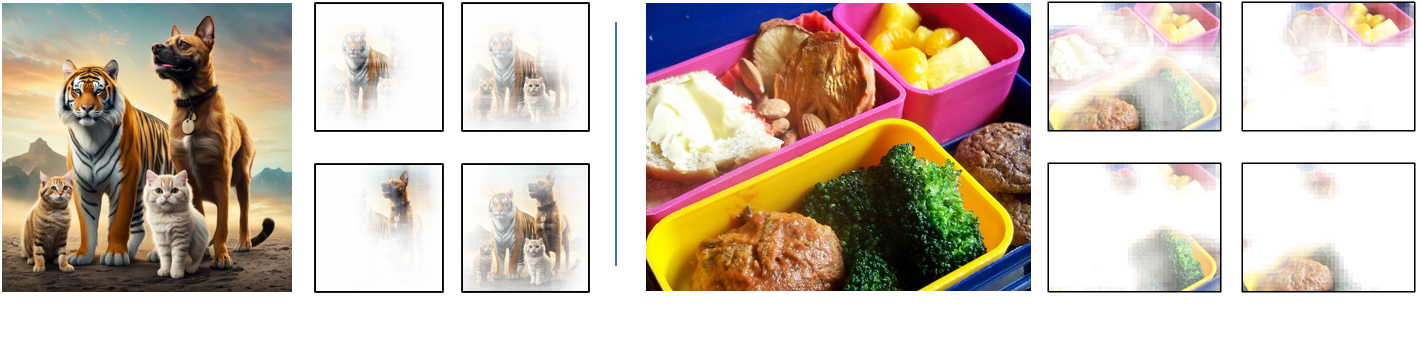}
        \put(-500,5){\large Original Image}
        \put(-393,68){\texttt{tiger}}
        \put(-344,68){\texttt{feline}}
        \put(-388,10){\texttt{dog}}
        \put(-342,10){\texttt{animal}}
        
        \put(-248,5){\large Original Image}
        \put(-117,68){\texttt{food}}
        \put(-47,68){\texttt{fruit}}
        \put(-130,10){\texttt{vegetable}}
        \put(-44,10){\texttt{meat}}
    }
    \caption{Example of Concept Visualizations, computed for a variety of WordNet synsets. We mask the image based on the saliency value. ConVis highlights the regions of the image that relate to any WordNet synset.
    }
    \label{fig:food_teaser}
\end{figure*}
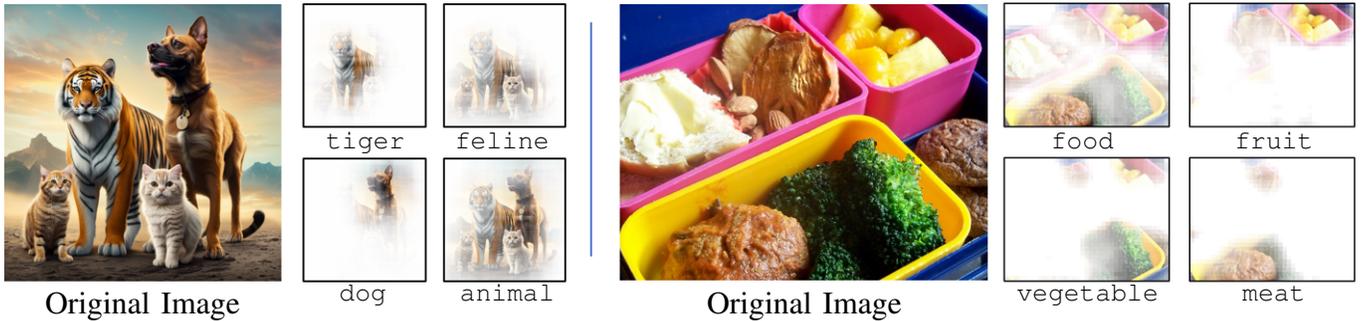

\section{Introduction}
\label{sec:intro}
The unstoppable progress of deep learning in Computer Vision (CV) has been largely driven by the development of pre-trained backbone networks~\cite{BackboneSurvey}. More recently, advancements in multi-modal foundation models such as CLIP~\cite{CLIP} have enabled breakthroughs in a variety of tasks, such as zero-shot classification~\cite{CLIP} and conditional image generation~\cite{DALLE2},~\cite{StableDiffusion}. 
Despite its impressive performance on a large number of tasks, CLIP relies on on complex black-box models, leading to lack of trust for models that employ CLIP in their pipeline.



By far the most popular solution to explain CV models are Saliency Maps~\cite{XAISurvey}.
However, given a classifier $\mathcal{M}$, existing saliency methodologies can only explain the model's predictions with respect to the classes pertaining to the model's task and training set. 
For models that employ backbones trained on a wider class range, much of the information encoded within the learned representations remains uninterpreted.
This is particularly relevant for CLIP, since \textit{i}) CLIP is trained on a very large and diverse dataset, and \textit{ii}) CLIP is trained on natural language supervision, without a training class set.

For example, consider a model $\mathcal{M}$ which utilizes the image embedding $\mathcal{E}^I$ from CLIP as a feature extraction network, and trained to classify \texttt{cat} and \texttt{dog} classes. Now, imagine presenting $\mathcal{M}$ with an image $\mathbf{x}$ featuring a \texttt{tiger}. In this scenario, $\mathcal{M}$ is likely to classify the \texttt{tiger} as a \texttt{cat}, since CLIP inherently captures the semantic proximity between \texttt{cat} and \texttt{tiger}, as both are \texttt{felines}. To explain this phenomenon, a Saliency Map needs to emphasize the semantic similarity, in CLIP space, between \texttt{cat} and \texttt{tiger}, as well as their connection to the overarching concept of \texttt{feline}. In this specific example, a Saliency Map for \texttt{feline} could explain the phenomenon by highlighting pixels belonging to both \texttt{cat} and \texttt{tiger}.
However, existing Saliency Maps are confined to interpreting the model's output solely in relation to the two training classes, \texttt{cat} and \texttt{dog}, hindering the ability to elucidate the relationship between \texttt{cat}, \texttt{tiger}, and the broader concept of \texttt{feline}.

To overcome this limitation, we propose Concept Visualization (ConVis): a novel saliency methodology that explains CLIP's image embedding.
To do so, we devise a similarity metric between images and semantic concepts sourced from the WordNet~\cite{WordNet} lexical database, taking into account the semantic relationships between concepts. This similarity is then computed for patches across a test image to be explained, thus obtaining local similarity scores, which are then aggregated to form a Saliency Map with respect to any WordNet synset $\mathbf{s}$.
ConVis is the first methodology able to provide Saliency Maps for \textit{any semantic concept}, also including concepts that are in principle not pertaining to the model's task or training set. 
Furthermore, ConVis is computed by using only information contained in CLIP's embedding space, and can thus explain CLIP \textit{agnostically to the end task}.


In Figure~\ref{fig:food_teaser}, we illustrate how ConVis enables the computation of Saliency Maps for any concept. These are computed independently of the classifier downstream from the CLIP backbone.
With regards to our previous example, from Figure~\ref{fig:food_teaser} it is clear how the semantic similarity between \texttt{cat} and \texttt{tiger} can be explained via the Saliency Map for \texttt{feline}, since the latter activates both on regions depicting \texttt{cat} and \texttt{tiger}. For an image $\mathbf{x}$ of a \texttt{tiger}, this would explain why a downstream \texttt{cat}/\texttt{dog} classifier would deem $\mathbf{x}$ to be a \texttt{cat}.

In our extensive experimental evaluation, we validate the efficacy of ConVis in explaining the CLIP embedding space. We verify the soundness of using information from WordNet in an Out Of Distribution (OOD) detection~\cite{OODSurvey} experiment. Furthermore, we compare to other popular saliency methodologies such as CAM~\cite{CAM} by computing ConVis on classes pertaining to the model's end task, demonstrating that ConVis achieves comparable performance. Lastly, we perform a user study to measure how well our methodology reflects how CLIP perceives the semantics of the input image, showing that users can guess the caption of a CLIP-embedded image only by looking at its explanations.  
Our contributions are:
\begin{itemize}
    \item We design a similarity metric in CLIP space between concepts in the WordNet~\cite{WordNet} lexical database and images.
    \item We propose a method employing this metric, multi-scale embedding, and spatial aggregation, to obtain Saliency Maps able to explain \textit{any semantic concept}.
\end{itemize}
\noindent Code: \href{https://github.com/loris2222/Concept-Visualization/}{https://github.com/loris2222/Concept-Visualization/}


\section{Related Work}
\label{sec:relwork}
Previous work in Explainable Artificial Intelligence~\cite{XAISurvey} have proposed methods to explain CV models. Among these, the most popular are
Saliency Maps. Given a model $\mathcal{M}$, an image $\mathbf{x}$, and a class $\mathbf{c}$, a Saliency Map is a heatmap of the same size of $\mathbf{x}$ that highlights the image regions that most contributed to the model's output with respect to class $\mathbf{c}$. 

Since the development of CAM~\cite{CAM}, many saliency methodologies have been developed. These are generally categorized in gradient-based and perturbation-based methods. Gradient-based methods, such as Grad-CAM~\cite{GradCAM}, compute saliency by analyzing the model's activations and gradients. Perturbation-based methods such as RISE~\cite{RISE}, instead, compute saliency by analyzing the model response to input perturbations.
While methods from both categories have seen widespread use, they are all limited to explaining predictions with respect to training classes, even when the model $\mathcal{M}$ to be explained employs backbones pre-trained on more diverse datasets. In the latter case, this limits explanations to a small portion of the encoded information. Conversely, ConVis can explain any concept.


Others have attempted to explain the latent semantics of deep models~\cite{SegDiscover, PV}, or producing inherently interpretable latent spaces~\cite{ConceptWhitening}, however, these require specific model architectures and cannot explain pre-trained models. Our proposed method, instead, can function with pre-trained CLIP models.



\section{Background}
\label{sec:background}
ConVis leverages the connection between text and images encoded within CLIP, integrating lexical information derived from WordNet.
In this section, we discuss relevant literature on pre-trained and foundation embedding networks
and present the WordNet lexical database.

\noindent\textbf{Embedding Networks: } 
In computer vision, convolutional backbones pre-trained on ImageNet~\cite{ImageNet} have been used in a variety of tasks, such as classification~\cite{AlexNet}, semantic segmentation~\cite{DeepLAB}, and object detection~\cite{Faster}. More recently, advances in multi-modal embeddings have enabled the encoding of inputs from different domains in a common space. In our work, we focus on explaining the CLIP~\cite{CLIP} multi-modal embedding, which jointly embeds text and images.

In its original formulation, CLIP is composed of a text embedding network $\mathcal{E}^T$ and of an image embedding network $\mathcal{E}^I$, each returning a vector $v \in \mathbb{R}^D$. These are jointly trained on a large dataset of captioned images belonging to a wide range of content. The training procedure follows the contrastive paradigm, such that the image embedding $\tilde{\mathbf{x}}_{i} = \mathcal{E}^I(\mathbf{x}_i)$ of image $\mathbf{x}_i$ is steered to be similar (in terms of cosine distance) to the text embedding $\tilde{c}_{i} = \mathcal{E}^T(c_i)$ of its caption $c_i$, and dissimilar to embeddings of other captions. 
The resulting embeddings reflect semantic similarity of the content of images and text, such that embeddings of images and text with similar semantic content are also similar in the feature space. In our work, we exploit this inherent semantic relationship to construct Saliency Maps.

\noindent\textbf{WordNet: }
WordNet~\cite{WordNet} is a large lexical database of the English language, structured as a graph where nodes represent \textit{synsets}, or sets of cognitive synonyms, and are linked by semantic and lexical relations, such as the ``is a" relation. For example, the concept \texttt{car} is in a ``is a" relation with the concept \texttt{motor\,\,vehicle}.

\begin{algorithm*}[ht]
    \begin{algorithmic}[1]
        \Require $\mathbf{x}$, $\mathbf{s}$, $\mathcal{E}^I$, $\mathcal{E}^T$, $\mathbb{T}$, $\delta_S$, $\delta_L$, $\omega$
        \For{$j=0$ to $H$ step $\omega$}
            \For{$i=0$ to $W$ step $\omega$}
                \vspace{2pt}
                \State $S_{\mathbf{x}}(i,j) = \mathbf{x}[j:j+\delta_S, i:i+\delta_S]$ \Comment{Compute patch at small scale}
                \vspace{3pt}
                \State $L_{\mathbf{x}}(i,j) = \mathbf{x}[j:j+\delta_L, i:i+\delta_L]$ \Comment{Compute patch at large scale}
                \vspace{2pt}
                \State $\tilde{\mathbf{e}}_{i,j} = \frac{\mathcal{E}^I(S_{\mathbf{x}}(i,j)) + \mathcal{E}^I(L_{\mathbf{x}}(i,j))}{2}$ \Comment{Compute mean of patch embeddings}
                \vspace{2pt}
                \State $\mathbf{r}_{i,j}(\mathbf{s}) = max\_rank\_sim(\tilde{\mathbf{e}}_{i,j}, \mathbf{s})$ \Comment{Compute the maximum rank similarity (Eq. \ref{eq:max_rank_sim})}
            \EndFor
        \EndFor
        \For{$i=0$ to $H$ step $\omega$}
            \For{$j=0$ to $W$ step $\omega$}
                \State $\mathbb{W}_{i,j} = \{\mathbf{r}_{l,m}(\mathbf{s}) \,{\boldsymbol{|}}\, |l-i| < \delta_L, |m-j| < \delta_L\}$ \Comment{Set of scores for patches that include pixel at $[i,j]$}
                \State $\mathbf{y}_\mathbf{x}^{\mathbf{s}}[i,j] = avg(\mathbb{W}_{i,j})$ \Comment{Compute Saliency Map pixel as the average of scores in $\mathbb{W}_{i,j}$}
            \EndFor
        \EndFor
        \vspace{3pt}
        \Return $\mathbf{y}_\mathbf{x}^{\mathbf{s}}$
    \end{algorithmic}
    \caption{Saliency map generation}
    \label{algo:procedure}
\end{algorithm*}

Each synset $\mathbf{s}$ in WordNet is identified by its unique name $name(\mathbf{s})$, and is associated to a set of phrases $lemmas(\mathbf{s})$ that encompass the concept's cognitive synonyms. 
Additionally, each synset is accompanied by a textual definition $\mathbf{d} = def(\mathbf{s})$ describing the concept. 
In our \texttt{car} example, the synset name is \texttt{car.n.01}, where \texttt{n} indicates that the concept is a noun, and \texttt{01} is an identifier used to distinguish between different meanings of \texttt{car}, such as \texttt{railway car}, which is identified by \texttt{car.n.02}.
The lemmas of \texttt{car.n.01} are \{\textit{car}, \textit{auto}, \textit{machine}, \textit{motorcar}\}, and its definition is ``\textit{a motor vehicle with four wheels; usually propelled by an internal combustion engine}".
In our work, we propose a  
CLIP-space synset-image similarity to obtain Saliency Maps.

\section{Concept Visualization}\label{sec:methods}
\begin{figure*}
    \centering
    \resizebox{0.78\linewidth}{!}{
        \begin{tikzpicture}
            \node[anchor=south east, inner sep=0] (image) at (0,0) {\includegraphics[width=\linewidth]{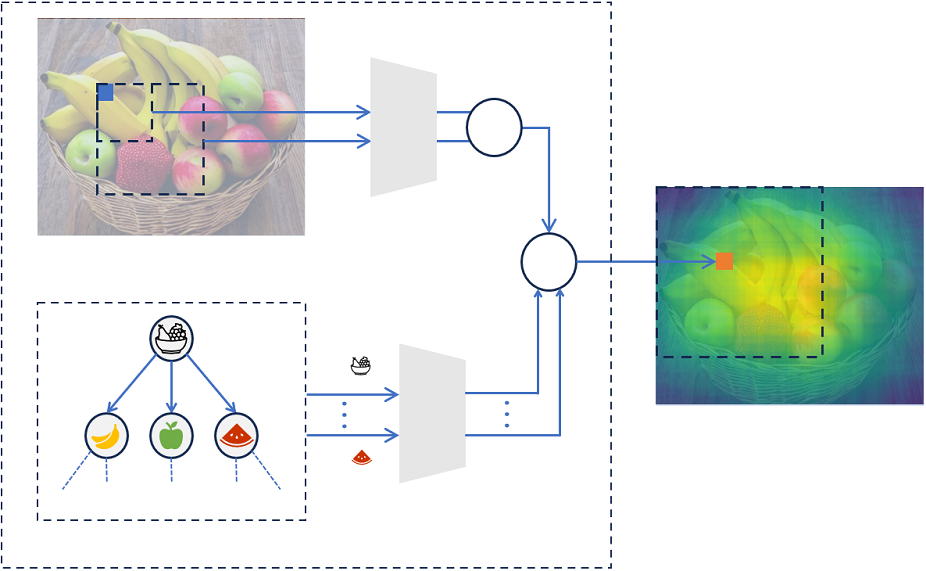}};
            
            \begin{scope}[x={(image.south west)},y={(image.north east)}]
                \fill[pattern=north west lines, pattern color=orange, opacity=0.7] (0.29, 0.67) rectangle (0.115, 0.375);
                \fill[color=white, opacity=0.7] (0.89, 0.36) rectangle (0.74, 0.32);
            \end{scope}
        \end{tikzpicture}
        \put(-456, 105){\large \texttt{fruit.n.01}}
        \put(-429,170){\huge $\mathbf{x}$}
        \put(-462,275){\large $\mathbf{x}[i,j]$}
        \put(-450,227){\large $\delta_S$}
        \put(-435,198){\large $\delta_L$}
        \put(-300,240){\huge $\mathcal{E}^I$}
        \put(-286,80){\huge $\mathcal{E}^T$}
        \put(-252,245){\Large $avg$}
        \put(-222,174){\Large max}
        \put(-220,163){\Large sim}
        \put(-80,75){\huge $\mathbf{y}$}
        \put(-120,182){\color{white}{\large $\mathbf{y}[i,j]$}}
        \put(-170,178){\large $\mathbf{r}_{i,j}$}
        \put(-344,111){\large $def(\,\,\,\,\,\,)$}
        \put(-344,60){\large $def(\,\,\,\,\,\,)$}
        \put(-135,105){\color{white}{\large $|m-j| < \delta_L$}}
        \put(-49,139){\rotatebox{90}{\color{white}{\large $ |l-i| < \delta_L$}}}
        \put(-498,15){\large $\{\mathbf{s} \in \mathbb{T} \,{\boldsymbol{|}}\, \mathbf{s} \,\text{is a}\, \text{\texttt{edible\,fruit}}\}$}
        \put(-390, 260){\large $S_{\mathbf{x}}(i,j)$}
        \put(-390, 244){\large $L_{\mathbf{x}}(i,j)$}
        \put(-203, 215){\large $\tilde{\mathbf{e}}_{i,j}$}
        \put(-490, 129){\huge $\mathbb{S}$}
        \put(-493, 297){\large \color{white} Input image}
        \put(-493, 33){\large WordNet hierarchy}
    }
    \caption{Concept Visualization computation diagram. The image patches at two different scales are encoded through CLIP's image embedding network $\mathcal{E}^I$, and concept definitions for synsets in $\mathbb{S}$ are encoded through CLIP's text embedding network $\mathcal{E}^T$. We then compute the $max\_rank\_sim$ and obtain scores $\mathbf{r}_{i,j}$ for each region in the image. The Saliency Map pixel $\mathbf{y}[i,j]$ is computed by averaging all scores $\mathbf{r}_{l,m}$ computed from patches that contained pixel $\mathbf{x}[i,j]$. The hatched rectangle on the right displays all the locations that satisfy this property. 
    }
    \label{fig:method_diagram}
\end{figure*}
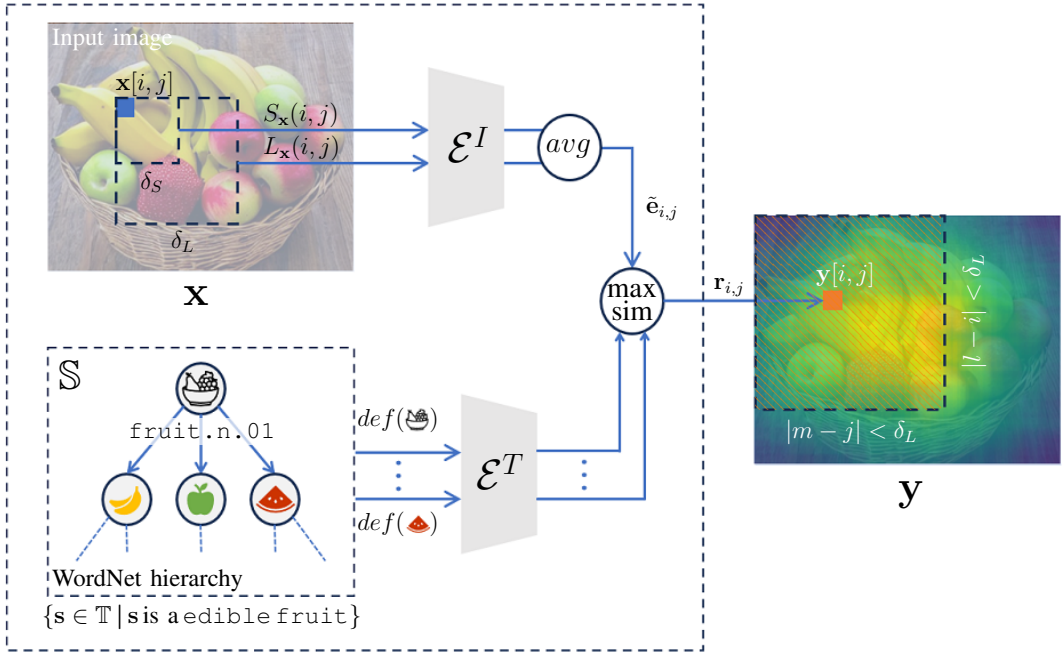

Given an image $\mathbf{x} \in \mathbb{R}^{W,H,C}$, any WordNet synset $\mathbf{s}$, and any model $\mathcal{M}$ employing CLIP as vision backbone, Concept Visualization returns a Saliency Map $\mathbf{y}_\mathbf{x}^{\mathbf{s}} \in \mathbb{R}^{W,H}$ that indicates the regions in $\mathbf{x}$ that, according to its CLIP embedding, most relate to the concept identified by $\mathbf{s}$.
Figure~\ref{fig:method_diagram} illustrates the computation of ConVis. First, all patches from the input image (top left) are encoded through CLIP's image encoder $\mathcal{E}^I$, and WordNet textual definitions for all concepts related to $\mathbf{s}$ (bottom left) are encoded using CLIP's text encoder $\mathcal{E}^T$. Image and text encodings are then compared using our proposed similarity metric (center) obtaining scores $\mathbf{r}_{i,j}$, which are finally spatially aggregated to compose the Saliency Map (right).
We now delineate the elements of this computation.


\noindent\textbf{Image-Concept Similarity: }
ConVis aims at highlighting the regions in the image that most relate to a semantic concept.
To this purpose, we 
define a similarity score between an image $\mathbf{x}$ and a synset $\mathbf{s} \in \mathbb{T}$, where $\mathbb{T}$ indicates the WordNet hierarchy, organized as a tree where the nodes represent synset and the edges represent the ``is a" relationship.
This score is then used to compute the values of the Saliency Map (Figure~\ref{fig:method_diagram}, center).

Similarity between any two CLIP embeddings $\tilde{\mathbf{e}}_1$, $\tilde{\mathbf{e}}_2$, regardless of their source modality (image or text), is computed via the cosine similarity $\phi(\tilde{\mathbf{e}}_1, \tilde{\mathbf{e}}_2)$. However, a synset is neither an image nor text, and as such it cannot be directly embedded in CLIP space. Thus, we leverage the lexical information contained in the synset's textual definition $\mathbf{d} = def(\mathbf{s})$ embedding it through $\mathcal{E}^T$. The rationale is that the definition $\mathbf{d}$ of a synset $\mathbf{s}$ entails its semantics, and as such its embedding $\tilde{\mathbf{d}} = \mathcal{E}^T(\mathbf{d})$ should be similar to the embedding $\tilde{\mathbf{x}} = \mathcal{E}^I(\mathbf{x})$ of an image $\mathbf{x}$ that visually displays that concept. We define a score to indicate this similarity between embeddings:
\begin{equation}
    \label{eq:z_score}
    z(\tilde{\mathbf{x}}, \mathbf{s}) = \phi(\tilde{\mathbf{x}}, \mathcal{E}^T(def(\mathbf{s}))).
\end{equation}

A limitation of Eq.~\ref{eq:z_score}, however, is that it calculates \textit{absolute} similarity, which can be influenced by the richness of visual details in the image. For instance, a portion of the image displaying a uniform blue color should ideally exhibit high similarity with the concept \texttt{sky}. Nonetheless, the scarcity of visual features may result in lower similarity scores for all synsets, including \texttt{sky.n.01}. To address this issue, we opt for a \textit{relative} approach that considers the similarity across all synsets in the WordNet hierarchy $\mathbb{T}$.
Given the 
synset $\mathbf{s}$ to be explained,
we define the rank similarity function as:
\begin{flalign}
\label{eq:rank_sim}
& rank\_sim(\tilde{\mathbf{x}}, \mathbf{s}) = \frac{|\mathbb{G}|}{|\mathbb{T}|}, && \\
&&& \mathmakebox[-55pt][r]{%
   \mathbb{G} = \{\mathbf{s}_{i} \in \mathbb{T} \,{\boldsymbol{|}}\, z(\tilde{\mathbf{x}}, \mathbf{s}_{i}) < z(\tilde{\mathbf{x}}, \mathbf{s})\},
  \hspace{-1.8em}
}
\notag
\end{flalign}
where $\mathbb{G}$ is the set of synsets in $\mathbb{T}$ that, according to $z$, are less similar to $\mathbf{x}$ than $\mathbf{s}$. The $rank\_sim$ is $0$ when $z(\tilde{\mathbf{x}}, \mathbf{s})$ is the lowest among all $z(\tilde{\mathbf{x}}, \mathbf{s}_i)$ for synsets $\mathbf{s}_i$ in $\mathbb{T}$, and is $1$ when it is the highest.

Until now, we have described similarity scores between an image $\mathbf{x}$ and a \textit{single} synset $\mathbf{s}$. Given a concept to be explained, however, the Saliency Map should highlight similarities between the image and \textit{any} synset that is in a ``is a" relationship with the synset that describes the concept. For example, the Saliency Map for concept \texttt{fruit}, identified by synset \texttt{edible\_fruit.n.01}, should also highlight regions of the image that are related to synsets such as \texttt{banana.n.02}.
To account for this, we first define the set $\mathbb{S}$ composed by all synsets in $\mathbb{T}$ that are in a ``is a" relationship with $\mathbf{s}$:
\begin{equation}
    \mathbb{S}_\mathbf{s} = \{\mathbf{s}_i \in \mathbb{T} \,{\boldsymbol{|}}\, \mathbf{s}_i \,\text{``\texttt{is\,a}"}\, \mathbf{s}\}.
\end{equation}
Then, we aggregate $rank\_sim$ over this set, as illustrated in Figure \ref{fig:method_diagram}, bottom left, and obtain:
\begin{equation}
    \label{eq:max_rank_sim}
    max\_rank\_sim(\tilde{\mathbf{x}}, \mathbf{s}) = \max_{\mathbf{s}_i \in \mathbb{S}_\mathbf{s}}(rank\_sim(\tilde{\mathbf{x}}, \mathbf{s}_i)).
\end{equation}
In Section \ref{sec:experiments}, we validate our decision to employ definitions as a source of lexical information, and we demonstrate the advantages of aggregating $rank\_sim$ over the set $\mathbb{S}_\mathbf{s}$.

\noindent\textbf{Saliency Map Generation: }
We have previously defined the $max\_rank\_sim$ score to measure similarity between visual concepts and synset $\mathbf{s}$  at the \textit{image level}. In the following, we detail the procedure to obtain \textit{pixel level} similarity, which is required to compute a Saliency Map. We follow the description of Algorithm \ref{algo:procedure} and Figure \ref{fig:method_diagram}.


Our idea is to compute local $max\_rank\_sim$ between the synset to be explained and image \textit{patches}.
To account for objects of different sizes, for each pixel location $[i,j]$ we crop small and large square patches $S_{\mathbf{x}}(i,j)$ and $L_{\mathbf{x}}(i,j)$ (lines \texttt{3-4}) (Figure \ref{fig:method_diagram}, top left):
\begin{flalign}
& \mathmakebox[0pt][l]{%
    \hspace{-8em}
    S_\mathbf{x}(i,j) = \mathbf{x}[j:j+\delta_S, i:i+\delta_S],
}
   \\
& \mathmakebox[0pt][l]{%
    \hspace{-8.05em}
    L_\mathbf{x}(i,j) = \mathbf{x}[j:j+\delta_L, i:i+\delta_L],
}
\notag
\end{flalign}
where $\delta_S$, $\delta_L$ indicate patch size.
We compute the embedding $\tilde{\mathbf{e}}_{i,j}$ representing the local image content as the average of the patches' embeddings (line \texttt{5}) (Figure \ref{fig:method_diagram}, top left):
\begin{equation}
    \tilde{\mathbf{e}}_{i,j} = \frac{\mathcal{E}^I(S_{\mathbf{x}}(i,j))+ \mathcal{E}^I(L_{\mathbf{x}}(i,j))}{2}.
\end{equation}
This operation is repeated for each valid location $[i,j]$ in the image, possibly taken at stride $\omega$ to expedite computation (lines \texttt{1-2}), obtaining local similarity scores $\mathbf{r}_{i,j}$ (line \texttt{6}):
\begin{equation}
    \mathbf{r}_{i,j}(\mathbf{s}) = max\_rank\_sim(\tilde{\mathbf{e}}_{i,j}, \mathbf{s}).
\end{equation}

After having computed the similarity scores $\mathbf{r}_{i,j}$ for every image region, we obtain the saliency values $\mathbf{y}_\mathbf{x}^{\mathbf{s}}[i,j]$ (lines \texttt{7-10}). Since the scores $\mathbf{r}_{i,j}$ are computed over overlapping image regions, the content of pixel $[i,j]$ in $\mathbf{x}$ contributes to multiple $\mathbf{r}_{i,j}$, as shown in the hatched rectangle in the right of Figure~\ref{fig:method_diagram}. 
Thus, to compute $\mathbf{y}_\mathbf{x}^{\mathbf{s}}[i,j]$, we first define the set $\mathbb{W}_{i,j}$ of all scores $\mathbf{r}_{l,m}$ that were affected by $\mathbf{x}[i,j]$ (line \texttt{9}):
\begin{equation}
    \mathbb{W}_{i,j} = \{\mathbf{r}_{l,m}(\mathbf{s}) \,{\boldsymbol{|}}\, |l-i| < \delta_L, |m-j| < \delta_L\}.
\end{equation}
The saliency value at each pixel location $[i, j]$ is then computed as the average of the scores in $\mathbb{W}_{i,j}$ (line \texttt{10}):
\begin{equation}
    \mathbf{y}_\mathbf{x}^{\mathbf{s}}[i,j] = avg(\mathbb{W}_{i,j}).
\end{equation}

\noindent\textbf{Constructing the Visualization: }
The value of $\mathbf{y}_\mathbf{x}^{\mathbf{s}}$ at each location $[i,j]$ reflects the semantic similarity between the concept identified by $\mathbf{s}$ and the corresponding image region.
To produce an explanation, the Saliency Map $\mathbf{y}_\mathbf{x}^{\mathbf{s}}$ can be color-coded using a color-map and be superimposed over $\mathbf{x}$, as in Figure \ref{fig:method_diagram}. Alternatively, $\mathbf{y}_\mathbf{x}^{\mathbf{s}}$ can be used as an interpolation factor between a fully white image and $\mathbf{x}$, such that the image is masked in regions where saliency is lower, and is visible in regions where saliency is higher, as shown in Figures~\ref{fig:food_teaser},~\ref{fig:app_images}.

\section{Experiments and Results}
\label{sec:experiments}

\begin{table*}
\caption{Results of OOD detection experiments on ImageNet subsets, chosen to encompass a wide variety of openness ($O^*$), cardinality ($|\boldsymbol{\cdot}|$), and semantic distance ($D$). 
We compare to MLS~\cite{MLS} and Isolation Forest~\cite{IFOR} over CLIP embeddings, and to CLIP max $img-img$ similarity over the dataset $\mathbf{Tr}^+$ (Eq. \ref{eq:clip_baseline}).}
\label{tab:ood}
\centering
\resizebox{\linewidth}{!}{
    \begin{tabular}{|c|lclccc|ccccc|}
\hline
\multicolumn{7}{|c|}{Dataset}                                                    & \multicolumn{5}{c|}{OOD AUROC} \\ \hline
\multicolumn{1}{|c}{} & $\Lambda^+$ & $|\Lambda^+|$                       & $\Lambda^-$ & $|\Lambda^-|$ & $O^*$ & $D$ & MLS          & IF & \makecell{CLIP \\ $img-img$} & \makecell{Ours \\ $rank$} & \makecell{Ours \\ $max\_rank$}         \\ \hline
1 & $\texttt{bird.n.01}$ & $59$ & $\texttt{hunting\_dog.n.01}$ & $63$           & $0.19$        & $6$                 & $0.970$         & $0.964$   & $0.994$ & $0.995$ & \boldmath{$0.997$}             \\
2 & $\texttt{garment.n.01}$ & $26$                                   & $\texttt{furniture.n.01}$ & $22$           & $0.16$        & $6$                 & \boldmath{$0.877$}         & $0.760$   & $0.740$ & $0.764$ & $0.862$             \\
3 & $\texttt{spider.n.01}$ & $6$                                   & $\texttt{butterfly.n.01}$ & $6$           & $0.18$        & $5$                 & $0.904$         & $0.953$   & $0.979$ & \boldmath{$0.995$} & \boldmath{$0.995$}             \\
4 & $\texttt{ungulate.n.01}$ & $17$                                  & $\texttt{bear.n.01}$ & $4$           & $0.05$        & $3$                 & $0.802$         & $0.914$   & $0.901$ & $0.743$ & \boldmath{$0.956$}             \\
5 & $\texttt{fruit.n.01}$ & $10$                                  & $\texttt{musical\_instrument.n.01}$ & $28$           & $0.35$       & $9$                & $0.989$         & $0.965$   & $0.991$ & $0.965$ & \boldmath{$0.993$}             \\
6 & $\texttt{aircraft.n.01}$ & $4$                                   & $\texttt{vessel.n.02}$ & $15$          & $0.41$       & $2$                & $0.735$         & $0.861$   & $0.935$ & $0.933$ & \boldmath{$0.939$}             \\ \hline
\multicolumn{5}{c}{} & \multicolumn{2}{|c|}{Average rank:} & $3.67$ & $3.83$ & $3.00$ & $3.00$ & \boldmath{$1.17$} \\ \cline{6-12}
\end{tabular}
}
\end{table*}

Our extensive experimental evaluation of Concept Visualization is organized as follows.
First, we demonstrate that embeddings of WordNet definitions correctly capture the concepts' semantics, we then compare ConVis to popular Saliency Maps via the WSOL benchmark proposed by Choe et al.~\cite{WSOLright}, and finally validate ConVis' efficacy through a user study.



\noindent\textbf{Experimental Setup: }
While ConVis can be applied to any synset, we filter WordNet and only consider the ``is a" ancestors of $1\,980$ physical objects from a hand-crafted list.
The result is a WordNet subset $\mathbb{T}$ encompassing $\sim7\,000$ synsets. Filtering was essential since WordNet's $117\,000$ synsets include abstract concepts like ``happiness", which are irrelevant for visual representation.


Experiments were carried out on a workstation with a single NVIDIA A6000 GPU. We explain OpenAI's publicly available \texttt{ViT-B/32} CLIP implementation. 
We set $\delta_S = 64$, $\delta_L = 128$, $\omega = 16$.
We chose this configuration  to obtain patches covering enough area for CLIP to perform well, but small enough that most visual concepts could be recognized. Indeed, CLIP does not perform well with images much smaller than those it was trained on, while too large patch sizes may result in small objects being ignored. 
We don't perform hyper-parameter tuning, but we hypothesize that the optimal patch size may depend on the dataset. 

Regarding computational complexity, given an image with size $(W,H)$ and stride $\omega$, our method requires to embed a number of patches $P \approx 2\times \frac{W \times H}{\omega^2}$, 
and then compute for each the $rank\_sim$ with at most $|\mathbb{T}|$ embeddings of WordNet definitions (pre-computed). In our setup ($W \times H \approx 300\,000$, $\omega = 16$), for the top-most concept in the hierarchy, the entire process takes $14$s on our hardware setup. This however is an extreme case, the time required is significantly lower on average. E.g. for the concept \texttt{dog}, the process takes $4$s.

\noindent\textbf{OOD Detection: } 
Concept Visualization operates under the assumption that a high $max\_rank\_sim$ between the embedding $\tilde{\mathbf{x}}$ and the synset $\mathbf{s}$ indicates that the image content aligns well with the semantic concept identified by $\mathbf{s}$. 
To validate this assumption, we use $max\_rank\_sim(\tilde{\mathbf{x}}, \mathbf{s})$ as an Out Of Distribution (OOD) score~\cite{OODSurvey} to determine whether an image $\mathbf{x}$ belongs to a set $\Lambda^+$ of known classes related to $\mathbf{s}$, or to an unknown class set $\Lambda^-$.
The underlying intuition is that if $max\_rank\_sim$ is a reliable OOD score, it follows that the proposed similarity metric is able to measure semantic similarity between images and WordNet synsets.


To measure OOD detection performance, we construct training dataset $\mathbf{Tr}^+ = \{\mathbf{x}_1 \compactdots \mathbf{x}_N\}$ of images pertaining to a set of known classes $\Lambda^+$, and test dataset $\mathbf{Te} = \mathbf{Te}^+ \cup \mathbf{Te}^-$ composed of images of classes in $\Lambda = \Lambda^+ \cup \Lambda^-$ including unknown classes $\Lambda^-$.
In this setup, the objective in the OOD detection task is to estimate the probability $o(\mathbf{x}_?)$ of a test image $\mathbf{x}_? \in \mathbf{Te}$ belonging to the set of known classes $\Lambda^+$. The chosen metric for evaluating OOD detection performance is the AUROC of $o$ over $\mathbf{Te}$.




In this experiment, our aim is to evaluate the efficacy of $max\_rank\_sim$ in assessing similarity between an image and a WordNet synset. To achieve this, we leverage the ImageNet dataset~\cite{ImageNet}, which stands out as an optimal choice for this investigation. Beyond its widespread use, ImageNet possesses a unique property: each class $\lambda$ in ImageNet's class set $\Omega$ is linked to a WordNet synset $\mathbf{s}_\lambda$. This association allows to construct the sets $\Lambda^+ = \{\lambda \in \Omega {\boldsymbol{|}} \mathbf{s}_\lambda \text{``\texttt{is a}"} \mathbf{s}^+\}$ and $\Lambda^- = \{\lambda \in \Omega {\boldsymbol{|}} \mathbf{s}_\lambda \text{``\texttt{is a}"} \mathbf{s}^-\}$ such that they comprise semantically related ImageNet classes, associated to synsets in a ``is a" relationship with $\mathbf{s}^+$ and $\mathbf{s}^-$, respectively.
By selecting these sets for the OOD test, we assess whether $o(\mathbf{x}_?)$ can discern similarities and differences between images $\mathbf{x}_?$ representing the synsets $\mathbf{s}^+$ and $\mathbf{s}^-$, even amongst their ``is a" descendants.
For example, given $s^+=\texttt{wildcat.n.03}$ and $s^-=\texttt{lemur.n.01}$, we can compute $\Lambda^+ = \{\text{cougar},\text{linx}\}$, $\Lambda^- = \{\text{indri, Madagascar cat}\}$. 

The sets $\mathbf{Tr}^+$, $\mathbf{Te}^+$, and $\mathbf{Te}^-$ are then constructed as ImageNet subsets, adhering to the original train/test split. 
The training set $\mathbf{Tr}^+$ comprises all ImageNet training images with label in $\Lambda^+$, while testing sets $\mathbf{Te}^+$ and $\mathbf{Te}^-$ consist of ImageNet test images with labels from $\Lambda^+$ and $\Lambda^-$, respectively.


To comprehensively evaluate OOD detection performance, we generate a variety of ImageNet subset triples $\mathbf{Tr}^+, \mathbf{Te}^+, \mathbf{Te}^-$ following the described approach. 
We select pairs of WordNet synsets $\mathbf{s}^+$, $\mathbf{s}^-$ to obtain $\Lambda^+, \Lambda^-$ pairs with varying characteristics. We examine \textit{cardinality} to understand the impact of the number of classes associated with each $\Lambda^+, \Lambda^-$ pair. Additionally, we consider \textit{openness} $O^*$, based on the ratio between in and out-of-distribution classes~\cite{OpenSetSurvey}, to inspect the effect of imbalance between $\Lambda^+$ and $\Lambda^-$. Lastly, we assess \textit{semantic distance} $D$, measured as the shortest path length in $\mathbb{T}$ between $\mathbf{s}^+$ and $\mathbf{s}^-$, to explore the behavior of score $o$ across different levels of separation between $\Lambda^+, \Lambda^-$. This approach enables a concise yet comprehensive examination of $max\_rank\_sim$ across diverse scenarios.

We test our proposed image-concept similarity by computing the performance of the OOD score $o(\mathbf{x}_?) = max\_rank\_sim(\tilde{\mathbf{x}}_?, \mathbf{s}^+)$, where $\tilde{\mathbf{x}}_?$ is the CLIP-encoded image $\tilde{\mathbf{x}}_? = \mathcal{E}^I(\mathbf{x}_?)$. Additionally, we perform an ablation study by computing the non-maximized $o(\mathbf{x}_?) = rank\_sim(\tilde{\mathbf{x}}_?, \mathbf{s}^+)$, to verify that aggregating over the set $\mathbb{S}$ described in Eq. \ref{eq:max_rank_sim} is beneficial to the computation of image-concept similarity. We also compare to a CLIP $img-img$ similarity baseline where the OOD score is defined as the maximum similarity (in CLIP space) between $\mathbf{x}_{?}$ and \textit{all} ImageNet training images belonging to known classes $\Lambda^+$:
\begin{equation}
\label{eq:clip_baseline}
    o(\mathbf{x}_?) = \max_{\mathbf{x} \in \mathbf{Tr}^+}(\phi(\mathcal{E}^I(\mathbf{x}), \mathcal{E}^I(\mathbf{x}_?))).
\end{equation}
The rationale is that images representing the same concept should be similar in CLIP space, and dissimilar if they represent different concepts.


We also compare to Maximum Logit Score (MLS)~\cite{MLS} and Isolation Forest (IF)~\cite{IFOR}, which are popular approaches to OOD detection. To ensure fair evaluation, we apply MLS and IF over CLIP embeddings of images in $\Lambda^+, \Lambda^-$ using the same CLIP model, which is a strong ImageNet classifier~\cite{CLIP}.

Our results (Table \ref{tab:ood}) show that $max\_rank\_sim$ is a good OOD score, as demonstrated by the superior performance of our image-definition similarity score with respect to the compared methods.
In particular, we note how the performance of MLS and IF drops when $\Lambda^+$ and $\Lambda^-$ have small semantic distance (row \texttt{6}), and for MLS even when the problem is made easier by a smaller openness (row \texttt{4}). Our approach, on the other hand, shows the best performance across all subsets except for the \texttt{garment.n.01} / \texttt{furniture.n.01} case, where it is nonetheless second only to MLS by a small margin. This is due to ImageNet images of garments also showing environments with furniture.

These results demonstrate that WordNet definitions are a good source of semantic information to compute Concept Visualizations, and that their CLIP embeddings align well with the CLIP embeddings of the visual concepts they describe.
Notably, the slightly better performance of $max\_rank\_sim$ compared to 
the CLIP image similarity baseline (Eq. \ref{eq:clip_baseline}) highlights the stronger alignment of WordNet definitions with the image's semantic content, even when compared to CLIP's $img-img$ similarity between images of the same classes.
Lastly, we demonstrate how $max\_rank\_similarity$ is a more robust similarity metric than its non-aggregated counterpart $rank\_similarity$, as demonstrated by its better performance across the board, but especially in row \texttt{4}.


\noindent\textbf{WSOL: } 
A common practice to quantitatively evaluate Saliency Maps is to measure their performance on GT-Known Weakly Supervised Object Localization (WSOL)~\cite{WSOLright}, which consists in predicting the bounding box of an object pertaining to a known class $\lambda$ within an image $\mathbf{x}$, without access to localization supervision. 
The intuition is that good Saliency Maps should focus on the object, thus, the localization performance of a Saliency Map reflects its explanation effectiveness.


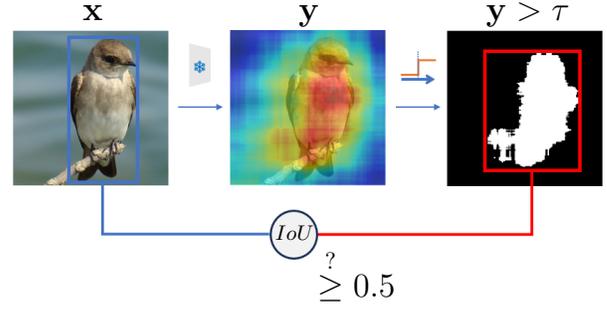
\begin{figure}[ht]
    \centering
    \resizebox{0.9\columnwidth}{!}{
    \includegraphics[width=\columnwidth]{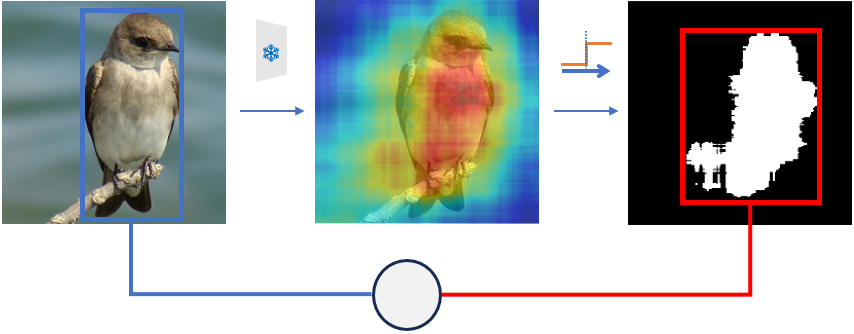}
    \put(-222,103){\Large $\mathbf{x}$}
    \put(-130,103){\Large $\mathbf{y}$}
    \put(-50,103){\Large $\mathbf{y}>\tau$}
    \put(-140,9){\small $IoU$}
    \put(-122,-15){\Large $\stackrel{?}{\geq}0.5$}
    }
    \caption{WSOL evaluation. We evaluate localization accuracy by computing the number of samples for which the $IoU$ between ground truth and Saliency Map is above $0.5$. To compute $IoU$, the Saliency Maps are thresholded at value $\tau$, which is optimized over the dataset.}
    \label{fig:WSOL}
\end{figure}

\begin{table}[t]
\caption{Results of the experiment on WSOL. We compare to popular CAM methodologies for CNNs.}
\label{tab:wsol}
\centering
\begin{tabular}{|l|c|}
\hline
\multicolumn{1}{|c|}{Method} & MaxBoxAcc     \\ \hline
Ours                  & \textit{76.2} \\ \hline
CAM~\cite{CAM}                          & 73.2          \\ \hline
HaS~\cite{HaS}                   & \textbf{78.1} \\ \hline
ACoL~\cite{ACoL}                         & 72.7          \\ \hline
SPG~\cite{SPG}                          & 71.4          \\ \hline
ADL~\cite{ADL}                          & 75.6          \\ \hline
CutMix~\cite{CutMix}                       & 71.9          \\ \hline
\end{tabular}
\end{table}

\begin{figure*}[ht]
    \centering
    \resizebox{0.7\linewidth}{!}{
        \includegraphics[width=\linewidth]{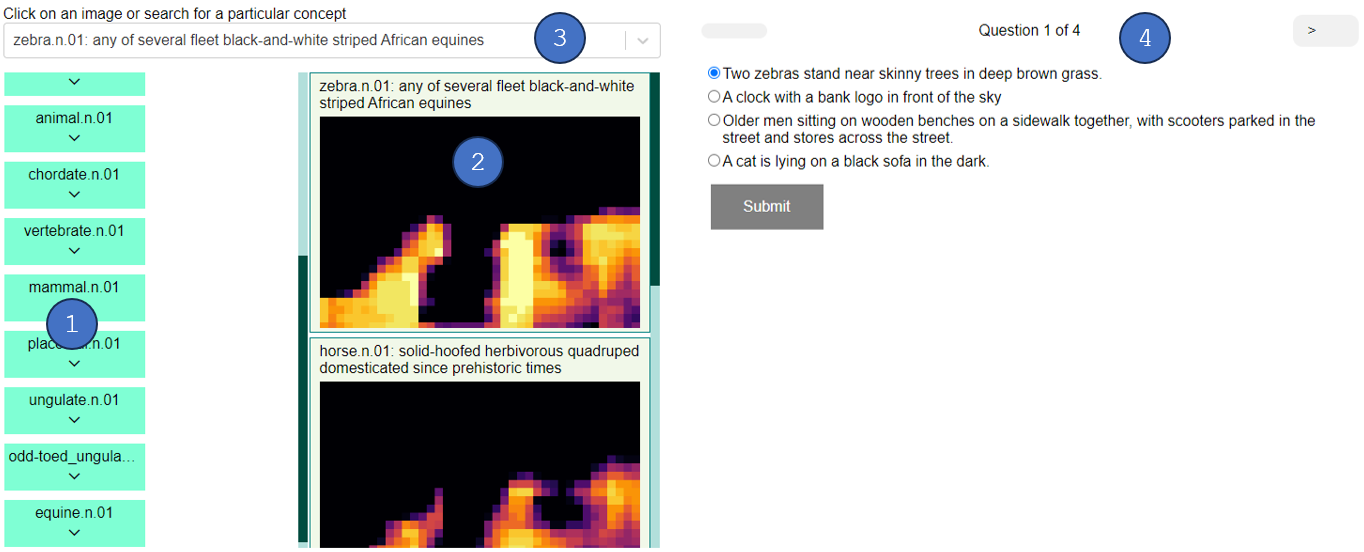}
    }
    \caption{Example question from the survey. \raisebox{.5pt}{\textcircled{\raisebox{-.9pt} {1}}} The user can explore Saliency Maps for concepts hierarchically, \raisebox{.5pt}{\textcircled{\raisebox{-.9pt} {2}}} can click on a Saliency Map of a concept to explore sub-concepts, \raisebox{.5pt}{\textcircled{\raisebox{-.9pt} {3}}} or search for specific concepts. \raisebox{.5pt}{\textcircled{\raisebox{-.9pt} {4}}} The user has to answer four questions, each one having one out of four correct answers in the form of COCO captions.}
    \label{fig:user_study_ui}
\end{figure*}

\begin{figure}[t]
    \begin{tikzpicture}
        \begin{axis}[
            ybar,
            width=\columnwidth,
            height=6cm,
            grid=both, 
            xmajorgrids=false, 
            ymajorgrids=true, 
            ymin=0,
            xtick=data,
            xtick style={draw=none},
            ytick={0,1,2,3,4,5,6,7,8},
            ytick style={draw=none},
            xticklabel style={font=\small},
            xlabel={Number of Correct Answers},
            ylabel={User Count},
            bar width=0.8cm,
            enlarge x limits=0.2,
            nodes near coords,
            every node near coord/.append style={font=\small, rotate=0, anchor=south},
            ]
            \addplot coordinates {(0, 0) (1, 1) (2, 3) (3, 8) (4, 6)};
        \end{axis}
    \end{tikzpicture}
    \caption{Results of the user study. We report the number of correct answers, out of four questions, for all 18 participants.
    }
    \vspace{-5pt}
    \label{fig:user_study_results}
\end{figure}
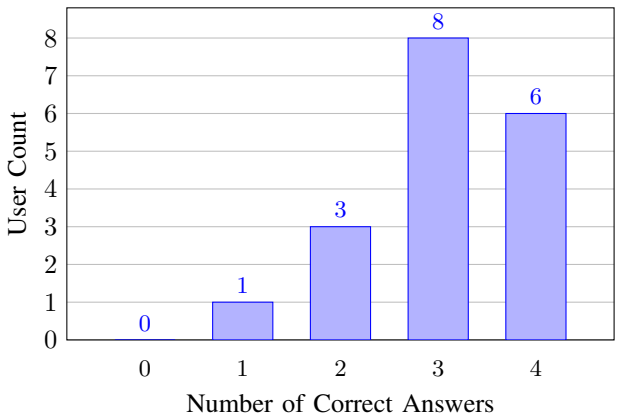

\begin{figure*}[ht]
    \centering
    \resizebox{0.86\linewidth}{!}{
        \includegraphics[width=\linewidth]{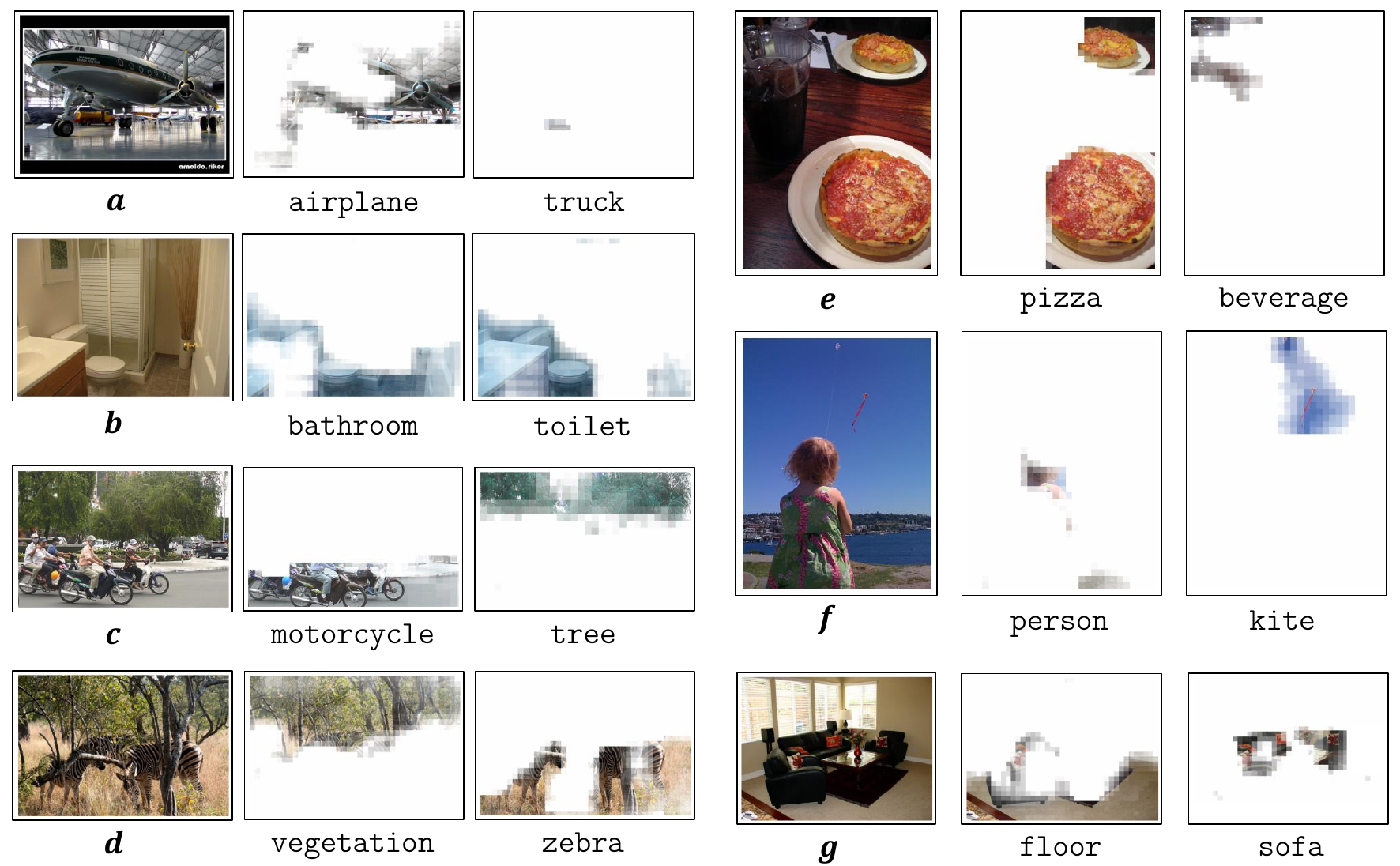}
    }
    \caption{
    ConVis applied to select COCO images from those employed in the user study, computed for visual concepts in the samples (participants to the user study had no access to images). 
    We highlight the limitations of recognizing small objects (sample $a$, \texttt{truck}), or objects with low contrast (sample $e$, \texttt{beverage}). Additionally, patches from central regions of sizable objects, like the plane in $a$, may lack context for accurate recognition. Sample $f$ emphasizes that only specific portions of an object may be considered similar to the concept, especially noticeable in instances related to \texttt{person}, where the model often identifies greater similarity with concepts such as \texttt{clothing}. Nonetheless, the explanations provided enough insight for the users to predict the captions only from the Saliency Maps.
    }
    \label{fig:app_images}
\end{figure*}

Following previous works~\cite{WSOLright}, we evaluate  WSOL performance 
by measuring Maximal Box Accuracy (\texttt{MaxBoxAcc}) over a dataset $\mathbf{X}$, defined as:

\vspace{-15pt}
\begin{flalign}
& \texttt{MaxBoxAcc} =  && \\
&&& \mathmakebox[0pt][r]{%
   \max_{\tau} \frac{1}{|\mathbf{X}|} \sum_{\mathbf{x} \in \mathbf{X}} \left[IoU(box(\mathbf{y}_\mathbf{x}^{\mathbf{c}}, \tau), B(\mathbf{x})) \geq \hat{\delta}\right],
}
\notag
\end{flalign}
where $box(\mathbf{y}_{\mathbf{x}}, \tau)$ is the bounding box of the largest connected component in Saliency Map $\mathbf{y}_{\mathbf{x}}^{\mathbf{c}}$ for class $\mathbf{c}$ after thresholding at $\tau$, and $B(\mathbf{x})$ is the ground truth bounding box for image $\mathbf{x}$. In practice, \texttt{MaxBoxAcc} measures the fraction of boxes extracted from Saliency Maps $\mathbf{y}_\mathbf{x}^{\mathbf{c}}, \mathbf{x} \in \mathbf{X}$ that match the ground truth boxes $B(\mathbf{x})$ with $IoU$ greater than $\hat{\delta}$, which is typically set to $0.5$ (Figure \ref{fig:WSOL}). The threshold $\tau$ is set as per~\cite{WSOLright}, by selecting the value $\tau \in [0, 1]$ that maximizes the fraction of correctly localized ($IoU \geq \hat{\delta}$) boxes.

We compare to mainstream Saliency Maps on the CUB fine-grained bird classification dataset~\cite{CUB}, which is a popular benchmark for WSOL evaluation~\cite{WSOLright}.
Results for~\cite{CAM, HaS, ACoL, SPG, ADL, CutMix} are as reported from~\cite{WSOLright}, and were obtained by computing the Saliency Maps for the ground truth class for a variety of CUB classifiers.
For each of the competing methods, we display the best result.
We compute ConVis without training a downstream CUB classifier, and compute saliency maps for $\mathbf{s} = \texttt{bird.n.01}$ instead of the specific bird species, as they might not be included in WordNet.
Since we are in a GT-Known scenario, however, competing methods are not penalized for classification errors, thus, it is fair to compare to a Saliency Map computed on the general concept \texttt{bird}.

Table \ref{tab:wsol} shows that the \texttt{MaxBoxAcc} of ConVis is in line with that of other Saliency Maps, with HaS~\cite{HaS} being the only method achieving slightly better performance. However, we stress that ConVis is the only methodology not being limited to the classes the model was trained on, and that it achieves this while matching the performance of other saliency methods, even when evaluated on classes the model was trained on.

\noindent\textbf{User Study: } 
So far, we have validated ConVis both at an image level, by measuring OOD detection performance, and at pixel level, by measuring WSOL performance. While these are useful, objective measures for Saliency Map quality, they may not reflect user experience. We thus perform a user study to validate ConVis' ability to provide insights on CLIP's perception of the image.


Previous attempts at evaluating Saliency Maps through user studies have introduced the concept of \textit{simulatability}~\cite{PV, PV1, PV8, PV12}, which measures the users' ability to predict the model's output \textit{from its explanations}.
In~\cite{PV, PV1, PV8, PV12}, simulatability was assessed by showing users \textit{both explanation and input image}, and asking them to guess the model's classification output.
The intuition is that for users to predict the model's output, the explanation must correctly capture and present the model's understanding of the input image. 

In a similar fashion, we measure simulatability with ConVis as explanation.
However, since ConVis explains the model's latent space agnostically to the end task, there is no classification output to be predicted. Instead, we measure simulatability by having users guess \textit{the caption} of the input image, which represents its semantic content. Clearly, this is trivial when the users have access to both explanation and input image, thus, in contrast with previous works, we ask users to guess captions by only seeing the Saliency Maps \textit{and not the images}, which is arguably much harder. 
If users can predict the caption of images by only seeing their explanations, it follows that ConVis is able to capture and convey enough information for them to understand the model's understanding of the input.

We structure the experiment as a gamified survey served via an interactive web application (Figure~\ref{fig:user_study_ui}). 
The interface enables to navigate Saliency Maps for all concepts in $\mathbb{T}$ hierarchically (Figure \ref{fig:user_study_ui}, \raisebox{.5pt}{\textcircled{\raisebox{-.9pt} {2}}}). For example, the user can click on the Saliency Map for \texttt{zebra.n.01} to see the explanations for its children concepts, such as \texttt{mountain\_zebra.n.01}, or click on ancestors to see explanations for more general concepts (Figure \ref{fig:user_study_ui}, \raisebox{.5pt}{\textcircled{\raisebox{-.9pt} {1}}}). Alternatively, users can search for any concept in $\mathbb{T}$ via the search bar and directly visualize the saliency map for the specified concept (Figure \ref{fig:user_study_ui}, \raisebox{.5pt}{\textcircled{\raisebox{-.9pt} {3}}}).

The user is tasked with guessing the caption of random images from the COCO~\cite{COCO} dataset. We choose COCO as it displays a diverse set of objects, rendering the task challenging for the user, and since it provides ground truth captions, which are needed for the experiment. Due to time and computational constraints on producing Saliency Maps for $\sim7\,000$ concepts per sample, we pre-compute Concept Visualizations for a set $U = \{\mathbf{x}_1, \dots, \mathbf{x}_{10}\}$ of $10$ random COCO samples (Figure~\ref{fig:app_images}), and for each user we randomly draw $4$ samples from $U$. For each sample, the user has to guess the ground truth caption amongst four possible options (Figure \ref{fig:user_study_ui}, \raisebox{.5pt}{\textcircled{\raisebox{-.9pt} {4}}}). 
To guarantee that the user may not exploit the caption's structure or wording, the three incorrect caption options are taken from ground truth COCO captions for samples outside of $U$, thus ensuring uniformity between the four options.

We gather responses from $18$ participants taken amongst Ms.C. and Ph.D. students with an AI background, obtaining $72$ total guesses. We report the results in Figure \ref{fig:user_study_results}. In particular, $1$ user guessed correctly only $1$ caption, $3$ users guessed $2$, $8$ users guessed $3$, and $6$ users guessed correctly all $4$ captions.

On average, the correct response rate was $76.4\%$, which is significantly higher than random chance. To validate this, we perform a one sample proportion binomial test. Our null hypothesis is $H_0: p=0.25$, which amounts to $25\%$ random chance of getting a correct answer amongst the four options. Given our sample size of $18$ and our resulting proportion of correct answers equal to $76.4\%$, we obtain p-value equal to $0.001$, with test statistic $X=13.8$. 
Thus, we reject $H_0$
and confirm that, thanks to ConVis, users can guess the caption of an image given only its explanation, doing so with a success rate much higher than random guessing. As such, we argue that our saliency methodology can help users gain insight on CLIP's perception of an image, and consequently gain insight on the functioning of models that employ CLIP as backbone.

\section{Conclusions and Future Works}
\label{sec:conclusions}
In this work, we have proposed Concept Visualization (ConVis), a novel saliency methodology that exploits lexical information from WordNet to explain the CLIP embedding, providing explanations that are agnostic to the end-task. In our experiments, we have validated ConVis on synthetic tasks, comparing it to other popular Saliency Maps. Furthermore, we have demonstrated the effectiveness of ConVis in a user study, showing that our methodology enables users to gain insight on CLIP's perception of the input image.


To the best of our knowledge, ConVis is the first saliency methodology able to explain the CLIP latent space for \textit{any} concept, including those concepts the end-model is not trained on. Consequently, ConVis is able to explain the entirety of the information encoded in CLIP's latent representation. We believe that using lexical information to compute Saliency Maps is a promising direction to obtain explanations for multi-modal backbones, and hope to inspire similar future works.



Future developments will explore the use of WordNet elements other than synset definitions, as well as different knowledge graphs. We also plan to apply ConVis to other VLP frameworks such as BLIP~\cite{BLIP}.
Furthermore, since our analysis was limited to physical objects, in the future we
may explore Saliency Maps focused towards more abstract concepts, potentially discovering interesting patterns in CLIP.

Lastly, we have formulated ConVis based on similarity between definitions and square image patches. Recent advancements such as the Segment Anything Network~\cite{SAM} could enable the use of custom-shaped regions, potentially improving the quality of the explanations.
Despite this, careful consideration should be given to the explainability of additional models in the pipeline.


\bibliographystyle{IEEEtran}
\bibliography{IJCNN/bibliography.bib}

\end{document}